# Neural Network Training with Asymmetric Crosspoint Elements

**Authors:** Murat Onen[1,2,*,†], Tayfun Gokmen[1,*,†], Teodor K. Todorov[1], Tomasz Nowicki[1], Jesús A. del Alamo[2], John Rozen[1], Wilfried Haensch[1], Seyoung Kim[1,*,‡]

**Affiliations:**

[1] IBM Thomas. J. Watson Research Center, Yorktown Heights, NY, 10598, USA.

[2] Massachusetts Institute of Technology, Cambridge, Massachusetts 02139, USA.

* Correspondence to: monen@mit.edu, tgokmen@us.ibm.com, kimseyoung@postech.ac.kr

† Authors contributed equally to this work.

‡ Current address: Department of Materials Science and Engineering, POSTECH, Pohang, Korea.

**Abstract:** Analog crossbar arrays comprising programmable nonvolatile resistors are under intense investigation for acceleration of deep neural network training. However, the ubiquitous asymmetric conductance modulation of practical resistive devices critically degrades the classification performance of networks trained with conventional algorithms. Here we first describe the fundamental reasons behind this incompatibility. Then, we explain the theoretical underpinnings of a novel fully-parallel training algorithm that is compatible with asymmetric crosspoint elements. By establishing a powerful analogy with classical mechanics, we explain how device asymmetry can be exploited as a useful feature for analog deep learning processors. Instead of conventionally tuning weights in the direction of the error function gradient, network parameters can be programmed to successfully minimize the total energy (Hamiltonian) of the system that incorporates the effects of device asymmetry. Our technique enables immediate realization of analog deep learning accelerators based on readily available device technologies.

**Main Text:**

Deep learning has caused a paradigm shift in domains such as object recognition, natural language processing, and bioinformatics which benefit from classifying and clustering representations of data at multiple levels of abstraction (*1*). However, the computational workloads to train state-of-the-art deep neural networks (DNNs) demand enormous computation time and energy costs for data centers (*2*). Since larger neural networks trained with bigger data sets generally provide better performance, this trend is expected to accelerate in the future. As a result, the necessity to provide fast and energy-efficient solutions for deep learning has invoked a massive collective research effort by industry and academia (*3–5*).

Highly optimized digital application-specific integrated circuit (ASIC) implementations have attempted to accelerate DNN workloads using reduced-precision arithmetic for the computationally intensive matrix operations. Although acceleration of inference tasks was achieved by using 2-bit resolution (*6*), learning tasks were found to require at least hybrid 8-bit floating-point formats (*7*) which still imposes considerable energy consumption and processing time for large networks. Therefore, beyond-digital approaches that can efficiently handle training workloads are actively sought for.

The concept of in-memory computation with analog resistive crossbar arrays is under intense study as a promising alternative. These frameworks were first designed to make use of Ohm's and Kirchhoff's Laws to perform parallel vector–matrix multiplications (See S2.1 an S2.2 for details), which constitute $\approx 2/3$ of the overall computational load (*8*). However, unless the remaining $\approx 1/3$ of computations during the update cycle is parallelized as well, the acceleration factors provided by analog arrays will be a mere $3 \times$ at best with respect to conventional digital processors. It was much later discovered that rank-one outer products can also be achieved in parallel, using pulse-coincidence and incremental changes in device conductance (*9*, *10*). Using this method, an entire crossbar array can be updated in parallel, without explicitly computing the outer product[1] or having to read the value of any individual crosspoint element (*11*). As a result, all basic primitives for DNN training using the Stochastic Gradient Descent (SGD) algorithm can be performed in a fully-parallel fashion using analog crossbar architectures. However, this parallel update method imposes stringent device requirements since its performance is critically affected by the conductance modulation characteristics of the crosspoint elements. In particular, asymmetric conductance modulation characteristics (i.e. having mismatch between positive and negative conductance adjustments) are found to deteriorate classification accuracy by causing inaccurate gradient accumulation (*9*, *11–16*). Unfortunately, all analog resistive devices to date have asymmetric characteristics, which poses a major technical barrier before the realization of analog deep learning processors.

In addition to widespread efforts to engineer ideal resistive devices (*17–20*), many high-level mitigation techniques have been proposed to remedy device asymmetry. Despite numerous published simulated and experimental demonstrations, none of these studies so far provides a solution for which the analog processor still achieves its original purpose: energy-efficient acceleration of deep learning. The critical issue with the existing techniques is the requirement of serial accessing to crosspoint elements one-by-one or row-by-row (*14–16*, *21–26*). Methods involving serial operations include reading conductance values individually, engineering update pulses to artificially force symmetric modulation, and carrying or resetting weights periodically. Furthermore, some approaches offload the gradient computation to digital processors, which not only requires consequent serial programming of the analog matrix, but also bears the cost of outer product calculation (*14*, *22–26*). Updating an $N \times N$ crossbar array with these serial routines would require at least $N$ or even $N^2$ operations. For practical array sizes, the update cycle would simply take too much computational time and energy. In conclusion, for implementations that compromise parallelism, whether or not the asymmetry issue is resolved becomes beside the point since computational throughput and energy efficiency benefits over conventional digital processors are lost for practical applications. It is therefore urgent to devise a method that deals with device asymmetry while employing only fully-parallel operations.

Recently, our group proposed a novel fully-parallel training method, *Tiki-Taka,* that can successfully train DNNs based on asymmetric resistive devices with asymmetric modulation characteristics (*27*). This algorithm was empirically shown in simulation to deliver ideal-device-equivalent classification accuracy for a variety of network types and sizes emulated with asymmetric device models (*27*). However, the missing theoretical underpinnings of the proposed algorithmic solution as well as the cost of doubling analog hardware previously limited the method described in Ref. (*27*).

---

[1] The result of the outer product is not returned to the user, but implicitly applied to the network.

In this paper, we first theoretically explain why device asymmetry has been a fundamental problem for SGD-based training. By establishing a powerful analogy with classical mechanics., we further establish that the *Tiki-Taka* algorithm minimizes the total energy (Hamiltonian) of the system, incorporating the effects of device asymmetry. The present work formalizes this new method as Stochastic Hamiltonian Descent (SHD) and describes how device asymmetry can be exploited as a useful feature in a fully-parallel training. The advanced physical intuition allows us to enhance the original algorithm and achieve a reduction in hardware cost of 50%, improving its practical relevance. Using simulated training results for different device families, we conclude that SHD provides better classification accuracy and faster convergence with respect to SGD-based training in all applicable scenarios. The contents of this paper provide a guideline for the next generation of crosspoint elements as well as specialized algorithms for analog computing.

**Theory**

Neural networks can be construed as many layers of matrices (i.e. weights, $W$) performing affine transformations followed by nonlinear activation functions. Training (i.e. learning) process refers to the adjustment of $W$ such that the network response to a given input produces the target output for a labeled dataset. The discrepancy between the network and target outputs is represented with a scalar error function, $E$, which the training algorithm seeks to minimize. In the case of the conventional SGD algorithm (*28*), values of $W$ are incrementally modified by taking small steps (scaled by the learning rate, $\eta$) in the direction of the gradient of the error function sampled for each input. Computation of the gradients is performed by the backpropagation algorithm consisting of forward pass, backward pass, and update subroutines (*29*) (**Fig. 1**A). When the discrete nature of DNN training is analyzed in the continuum limit, the time evolution of $W$ can be written as a Langevin equation:

$$\dot{W} = -\eta \left[ \frac{\partial E}{\partial W} + \epsilon(t) \right] \qquad (1)$$

where $\eta$ is the learning rate and $\epsilon(t)$ is a fluctuating term with zero-mean, accounting for the inherent stochasticity of the training procedure (*30*). As a result of this training process, $W$ converges to the vicinity of an optimum $W_0$, at which $\frac{\partial E}{\partial W} = 0$ but $\dot{W}$ is only on average 0 due to the presence of $\epsilon(t)$. For visualization, if the training dataset is a cluster of points in space, $W_0$ is the center of that cluster, where each individual point still exerts a force ($\epsilon(t)$) that averages out to 0 over the whole dataset.

In the case of analog crossbar-based architectures, the linear matrix operations are performed on arrays of physical devices, whereas all nonlinear computations (e.g. activation and error functions) are handled at peripheral circuitry. The strictly positive nature of device conductance requires representation of each weight by means of the differential conductance of a pair of crosspoint elements (i.e. $W \propto G_{main} - G_{ref}$). Consequently, vector-matrix multiplications for the forward and backward passes are computed by using both the main and the reference arrays (**Fig.1**A). On the other hand, the gradient accumulation and updates are only performed on the main array using bidirectional conductance changes while the values of the reference array are kept constant[2]. In this section, to illustrate the basic dynamics of DNN training with analog architectures, we study a single-parameter optimization problem (linear regression) which can be considered as the simplest "neural network".

The weight updates in analog implementations are carried out through modulation of the conductance values of the crosspoint elements, which are often applied by means of pulses. These pulses cause incremental changes in device conductance ($\Delta G^{+,-}$). In an ideal device, these modulation increments are of equal magnitude in both directions and independent of the device conductance, as shown in **Fig. 1**B. It should be noted that the series of modulations in the training process is inherently non-monotonic as different input samples in the training set create gradients with different magnitudes and signs in general. Furthermore, as stated above, even when an optimum conductance, $G_0$, is reached ($W_0 \propto G_0 - G_{ref}$), continuing the training operation would continue modifying the conductance in the vicinity of $G_0$, as shown in **Fig. 1**C. Consequently, $G_0$ can be considered as a dynamic equilibrium point of the device conductance from the training algorithm point of view.

Despite considerable technological efforts in the last decade, analog resistive devices with the ideal characteristics illustrated in **Fig. 1**B have yet to be realized. Instead, practical analog resistive devices display asymmetric conductance modulation characteristics such that unitary (i.e. single-pulse) modulations in opposite directions do not cancel each other in general, i.e., $\Delta G^+(G) \neq -\Delta G^-(G)$. However, with the exception of some device technologies such as Phase Change Memory (PCM) which reset abruptly (*15, 21, 31*), many crosspoint elements can be modeled by a smooth, monotonic, nonlinear function that shows saturating behavior at its extrema as shown in **Fig. 1**E (*19, 32, 33*). For such devices, there exists a unique conductance point, $G_{symmetry}$, at which the magnitude of an incremental conductance change is equal to that of a decremental one. As a result, the time evolution of $G$ during training can be rewritten as:

$$\dot{G} = -\eta \left[ \frac{\partial E}{\partial G} + \epsilon(t) \right] - \eta \kappa \left| \frac{\partial E}{\partial G} + \epsilon(t) \right| \cdot f_{hardware} \quad (2)$$

where $\kappa$ is the asymmetry factor and $f_{hardware}$ is the functional form of the device asymmetry (See S1.1 for derivation). In this expression, the term $-\eta \left| \frac{\partial E}{\partial G} + \epsilon(t) \right|$ signifies that the direction of the change related to asymmetric behavior is solely determined by $f_{hardware}$, irrespective of the direction of the intended modulation. For the exponentially saturating device model shown in **Fig.1**E, $f_{hardware} = G - G_{symmetry}$, which indicates that each and every update event has a component that drifts the device conductance towards its symmetry point. A simple observation of this effect is when enough equal number of incremental and decremental changes are applied to these devices in a random order, the conductance value converges to the vicinity of $G_{symmetry}$ (*33*). Therefore, this point can be viewed as the physical equilibrium point for the device, as it is the only conductance value that is dynamically stable.

---

[2] For implementations using devices showing unidirectional conductance modulation characteristics, both the main and the reference array are updated. When SGD is used as the training algorithm, values of $G_{ref}$ are not critical as long as they fall in the midrange of $G_{main}$'s conductance span (*9*).

It is essential to realize that there is in general no relation between $G_{symmetry}$ and $G_0$, as the former is entirely device-dependent while the latter is problem-dependent. As a result, for an asymmetric device, two equilibria of hardware and software create a competing system, such that the conductance value converges to a particular conductance somewhere between $G_{symmetry}$ and $G_0$, for which the driving forces of the training algorithm and device asymmetry are balanced out. (**Fig.1**F). In examples shown in **Fig.1**C and **1**F, $G_0$ of the problem is purposefully designed to be far away from $G_{symmetry}$, so as to depict a case for which the effect of asymmetry is pronounced. Indeed, it can be seen that the discrepancy between the final converged value, $G_{final}$, and $G_0$ strongly depends on the relative position of $G_0$ with respect to the $G_{symmetry}$ (**Fig.1**G), unlike that of ideal devices (**Fig.1**D). Detailed derivation of these dynamics can be found in S1.2.

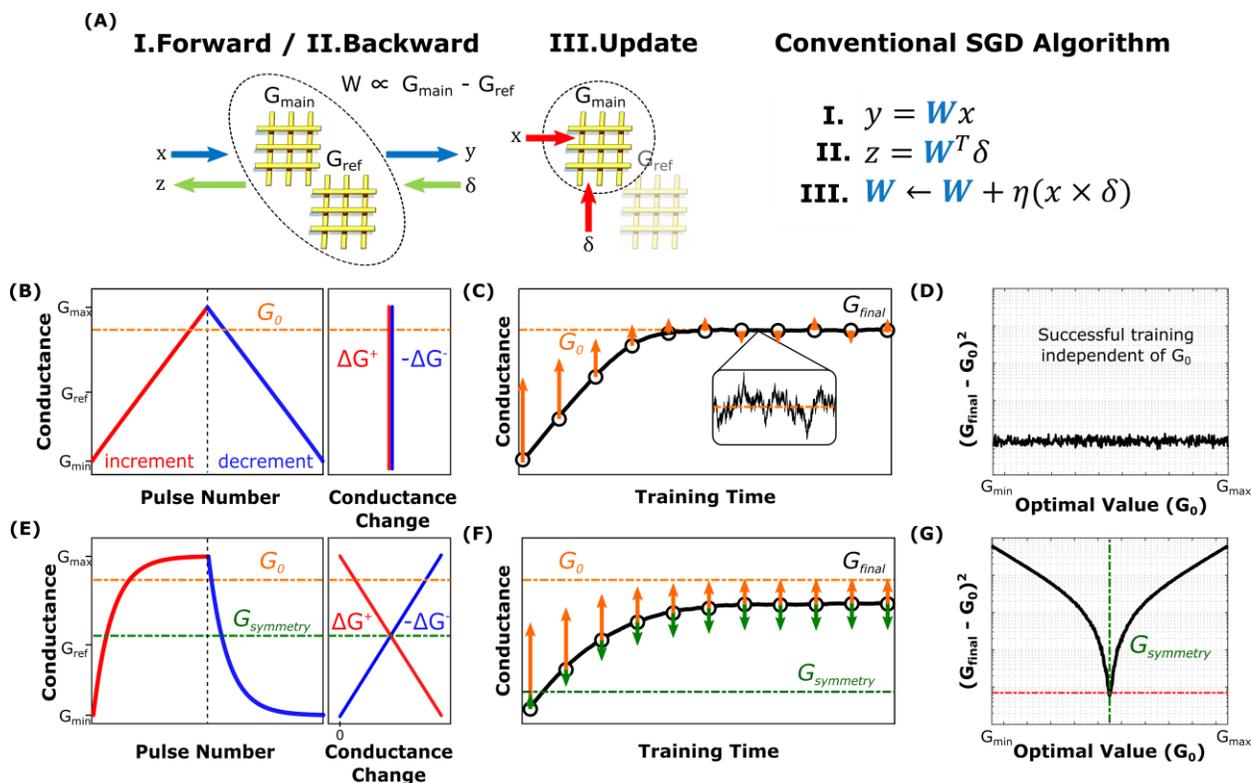

**Fig.1 Effect of asymmetric conductance modulation for SGD-based training.** (**A**) Schematic and pseudocode of processes for conventional SGD algorithm (*28*). Vectors $x, y$, represent the input and output vectors in the forward pass whereas $\delta, z$ contain the backpropagated error information. The analog architecture schematic is only shown for a single layer, where all vectors are propagated between upper and lower network layers in general. The pseudocode only describes operations computed in the analog domain, whereas digital computations such as activation functions are not shown for simplicity. (**B**) Sketch of conductance modulation behavior of a symmetric crosspoint device. (**C**) Simulated single-

parameter optimization result for the symmetric device shown in (B). conductance successfully converges to the optimal value for the problem at hand, $G_0$. **(D)** Simulated residual distance between the final converged value, $G_{final}$, and $G_0$ for training the device with characteristics shown in (B) for datasets with different optimal values. **(E)** Sketch of conductance modulation behavior of an asymmetric crosspoint device. The point at which $\Delta G^+ = \Delta G^-$ is defined as the symmetry point of the device ($G_{symmetry}$) **(F)** Simulated training result for the same single-parameter optimization with the asymmetric device shown in (E). Device conductance fails to converge to $G_0$, but instead settles at a level between $G_0$ and $G_{symmetry}$. **(G)** Simulated residual distance (in semilog scale) between the final value, $G_{final}$, and $G_0$ for training the device with characteristics shown in (E) for datasets with different optimal values. All simulation details can be found in S2.5.

In contrast to SGD, our new training algorithm, illustrated in **Fig.2**A, separates both the forward path and error backpropagation from the update function. For this purpose, two array pairs (instead of a single pair), namely $A_{main}, A_{ref}, C_{main}, C_{ref}$ are utilized to represent each layer (*27*). In this representation, $A = A_{main} - A_{ref}$ stands for the auxiliary array and $C = C_{main} - C_{ref}$ stands for the core array.

The new training algorithm operates as follows. At the beginning of the training process, $A_{ref}$ and $C_{ref}$ are initialized to $A_{main,symmetry}$ and $C_{main,symmetry}$, respectively (reasons will be clarified later, detailed method can be found in S2.3), following the method described in Ref.(*33*). As illustrated in **Fig.2**A, first, forward and backward pass cycles are performed on the array-pair $C$ (Steps *I* and *II*), and corresponding updates are performed on $A_{main}$ (scaled by the learning rate $\eta_A$) using the parallel update scheme discussed in Ref. (*9*) (Step *III*). In other words, the updates that would have been applied to $C$ in a conventional SGD scheme are directed to $A$ instead.

Then, every $\tau$ cycles, another forward pass is performed on $A$, with a vector $u$, which produces $v = Au$ (Step *IV*). In its simplest form, $u$ can be a vector of all "0"s but one "1", which then makes $v$ equal to the row of $A$ corresponding to the location of "1" in $u$. Finally, the vectors $u$ and $v$ are used to update $C_{main}$ with the same parallel update scheme (scaled by the learning rate $\eta_c$) (Step *V*). These steps (*IV* and *V* shown in **Fig 2.**A) essentially partially add the information stored in $A$ to $C_{main}$. The complete pseudocode for the algorithm can be found in S2.4.

At the end of the training procedure $C$ alone contains the optimized network, to be later used in inference operations (hence the name core). Since A receives updates computed over $\frac{\partial E}{\partial C}$, which have zero-mean once $C$ is optimized, its active component, $A_{main}$, will be driven towards $A_{main,symmetry}$. The choice to initialize the stationary reference array, $A_{ref}$, at $A_{main,symmetry}$ ensures that $A = 0$ at this point (i.e. when $C$ is optimized), thus generating no updates to $C$ in return.

With the choice of $u$ vectors made above, every time steps *IV* and *V* are performed, the location of the "1" for the $u$ vector would change in a cyclic fashion, whereas in general any set of orthogonal $u$ vectors can be used for this purpose (*27*). We emphasize that these steps should not be confused with weight carrying (*15*, *16*), as $C$ is updated by only a fractional amount in the direction of $A$ as $\eta_C \ll 1$ and at no point information stored in $A$ is externally erased (i.e. $A$ is never reset). Instead, $A$ and $C$ create a coupled-dynamical-system, as the changes performed on both are determined by the values of one another.

Furthermore, it is critical to realize that the algorithm shown in **Fig. 2** consists of only fully-parallel operations. Similar to steps *I* and *II* (forward and backward pass on $C$), steps *IV* is yet another matrix-vector multiplication that is performed by means of Ohm's and Kirchhoff's Laws. On the other hand, the update steps *III* and *V* are performed by the stochastic update scheme (*9*). This update method does not explicitly compute the outer products ($x \times \delta$ and $u \times v$), but instead uses a statistical method to modify all weights in parallel proportional to those outer products. As a result, no serial operations are required at any point throughout the training operation, enabling high throughput and energy efficiency benefits in deep learning computations.

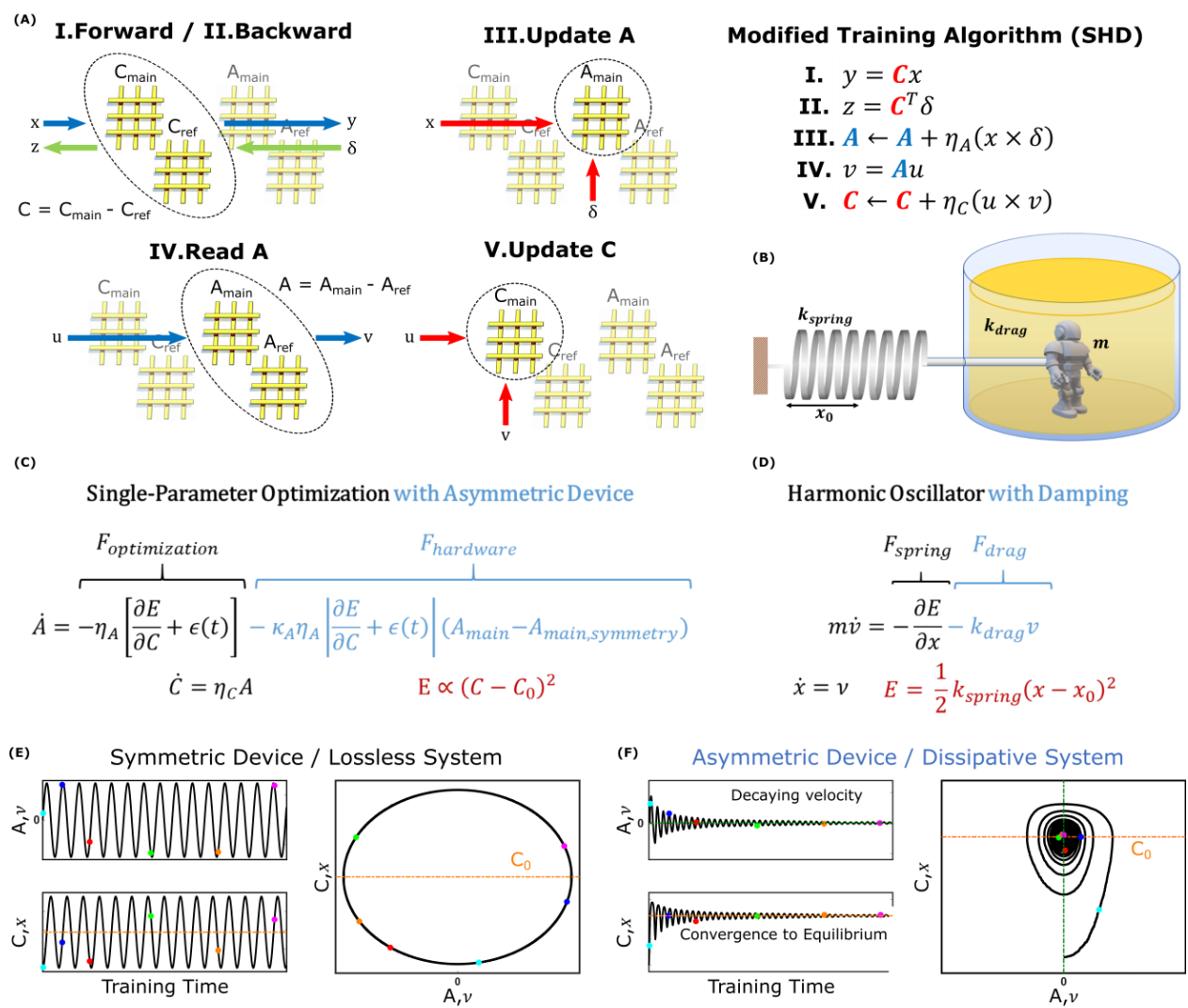

**Fig.2 DNN training with Stochastic Hamiltonian Descent (SHD) algorithm and dynamics of a**

dissipative harmonic oscillator. **(A)** Schematic and pseudocode of training process using the SHD algorithm. The pseudocode only describes operations computed in the analog domain, whereas digital computations such as nonlinear error functions are not shown for simplicity. **(B)** Illustration of a damped harmonic oscillator system. **(C)** Differential equations describing the evolution of the parameters with the SHD training algorithm in the continuum limit. **(D)** Equations of motion describing the dynamics of a harmonic oscillator. **(E)** Simulated results for a single-parameter optimization task using the SHD algorithm with symmetric devices described in **Fig.1**B. **(F)** Simulated results for a single-parameter optimization task using the SHD algorithm with asymmetric devices described in **Fig.1**E. All simulation details can be found in S2.5.

For the same linear regression problem studied above, the discrete-time update rules given in **Fig.2**A can be rewritten as a pair of differential equations in the continuum limit that describe the time evolution of subsystems $A$ and $C$ (**Fig.2**C) as:

$$\dot{A} = -\eta_A \left[ \frac{\partial E}{\partial C} + \epsilon(t) \right] - \eta_A \kappa_A \left| \frac{\partial E}{\partial C} + \epsilon(t) \right| (A_{main} - A_{main,\,symmetry}) \tag{3}$$

$$\dot{C} = \eta_C A + \eta_C \kappa_C |A| (C_{main} - C_{main,symmetry}) \tag{4}$$

It can be noticed that this description of the coupled system has the same arrangement as the equations governing the motion of a damped harmonic oscillator (**Fig.2**B, D). In this analogy, subsystem $A$ corresponds to velocity, $v$, while subsystem $C$ maps to position, $x$, allowing the scalar error function of the optimization problem[3], $(C - C_0)^2$, to map onto the scalar potential energy of the physical framework, $\frac{1}{2} k_{spring}(x - x_0)^2$. Moreover, for implementations with asymmetric devices, an additional force term, $F_{hardware}$, needs to be included in the differential equations to reflect the hardware-induced effects on the conductance modulation. As discussed earlier, for the device model shown in **Fig.1**E this term is proportional to $A_{main} - A_{main,symmetry}$. If we assume $A_{ref} = A_{main,symmetry}$ (this assumption will be explained later), we can rewrite $F_{hardware}$ as a function of $A_{main} - A_{ref}$, which then resembles a drag force, $F_{drag}$, that is linearly proportional to velocity ($v \propto A = A_{main} - A_{ref}$) with a variable (but strictly nonnegative) drag coefficient $k_{drag}$. In general, the $F_{hardware}$ term can have various functional forms for devices with different conductance modulation characteristics but is completely absent for ideal devices. Note that, only to simplify the physical analogy, we ignore the effect of asymmetry in subsystem $C$, which yields the equation shown in Fig.2C (instead of Eq. 4). This decision will be justified in the Discussions section and is derived in detail in S1.3.

Analogous to the motion of a lossless harmonic oscillator, the steady-state solution for this modified optimization problem with ideal devices (i.e. $F_{hardware} = 0$) has an oscillatory behavior **(Fig 2**E). This result is expected, as in the absence of any dissipation mechanism, the total energy of the system cannot be minimized (it is constant) but can only be continuously transformed between its potential and kinetic components. On the other hand, for asymmetric devices, the dissipative force term $F_{hardware}$ gradually annihilates all energy in the system, allowing $A \propto v$ to converge to 0 ($E_{kinetic} \to 0$) while $C \propto x$ converges to $C_0 \propto x_0$ ($E_{potential} \to 0$). Based on these observations, we rename the new training algorithm as *Stochastic Hamiltonian Descent (SHD)* to highlight the evolution of the system parameters in the direction of reducing the system's total energy (Hamiltonian). These dynamics can be visualized by plotting the time evolution of $A$ versus that of $C$, which yields a spiraling path representing decaying oscillations for the optimization process with asymmetric devices (**Fig 2**F), in contrast to elliptical trajectories observed for ideal lossless systems (**Fig 2**E).

---

[3] Conventionally error functions are written in terms of the difference between the network response and the target output and gradients are computed accordingly. However, in the absence of any stochasticity, $\epsilon$, it can instead be written in terms of the network weights and their optimal values as well for notational purposes.

Following the establishment of the necessity to have dissipative characteristics, here we analyze conditions at which device asymmetry provides this behavior. It is well-understood in mechanics that for a force to be considered dissipative, its product with velocity (i.e. power) should be negative (otherwise it would imply energy injection into the system). In other words, the hardware-induced force term $F_{hardware} = -\kappa_A \eta_A \left| \frac{\partial E}{\partial C} + \epsilon(t) \right| (A_{main} - A_{main,symmetry})$ and the velocity, $v = A_{main} - A_{ref}$, should always have opposite signs. Furthermore, from the steady-state analysis, for the system to be stationary ($v = 0$) at the point with minimum potential energy ($x = x_0$), there should be no net force ($F = 0$). Both of these arguments indicate that, for the SHD algorithm to function properly, $A_{ref}$ must be set to $A_{main,symmetry}$. Note that as long as the crosspoint elements are realized with asymmetric devices (opposite to SGD requirement) and a symmetry point exists for each device, the shape of their modulation characteristics is not critical for successful DNN training with the SHD algorithm. Importantly, while a technologically viable solution for symmetric devices has not yet been found over decades of investigation, asymmetric devices that satisfy the aforementioned properties are abundant. To validate this claim, we present an experimental demonstration these dynamics using metal-oxide based electrochemical devices (*32*) in S3.1 (**Fig.S3**).

A critical aspect to note is that the SGD and the SHD algorithms are fundamentally disjunct methods governed by completely different dynamics. The SGD algorithm attempts to optimize the system parameters while disregarding the effect of device asymmetry and thus converges to the minimum of a wrong energy function. On the other, the system variables in an SHD-based training do not conventionally evolve in directions of the error function gradient, but instead, are tuned to minimize the total energy incorporating the hardware-induced terms. The most obvious manifestation of these properties can be observed when the training is initialized from the optimal point (i.e. the very lucky guess scenario) since any "training" algorithm should at least be able to maintain this optimal state. For the conventional SGD, when $W = W_0$, the zero-mean updates applied to the network were shown above to drift $W$ away from $W_0$ towards $W_{symmetry}$. On the other hand, for the SHD method, when $A = 0$ and $C = C_0$, the zero-mean updates applied on $A$ do not have any adverse effect since $A_{main}$ is already at $A_{main,symmetry}$ for $A = 0$. Consequently, no updates are applied to $C$ either as $\dot{C} = A = 0$. Therefore, it is clear that SGD is fundamentally incompatible with asymmetric devices, even when the solution is guessed correctly from the beginning, whereas the SHD does not suffer from this problem (See S1.2 and S1.3). Note that the propositions made for SGD can be further generalized to other crossbar-compatible training methods such as equilibrium propagation (*34*) and deep Boltzmann machines (*35*), which can also be adapted to be used with asymmetric devices following the approach discussed in this paper. Additional discussions on SHD operation can be found in S3.3 and S3.4.

Finally, we appreciate that large-scale neural networks are much more complicated systems with respect to the problem analyzed here. Similarly, different analog devices show a wide range of conductance modulation behaviors, as well as bearing other non-idealities such as analog noise, imperfect retention, and limited endurance. However, the theory we provide here finally provides an intuitive explanation for: (1) why device asymmetry is fundamentally incompatible with SGD-based training and (2) how to ensure accurate optimization while only using fully-parallel operations. We conclude that asymmetry-related issues within SGD should be analyzed in the context of competing equilibria, where the optimum for the classification problem is not even a stable solution at steady-state. In addition to this simple stability analysis, the insight to modify the optimization landscape to include nonideal hardware effects allows other fully-parallel solutions to be designed in the future using advanced concepts from optimal control theory. As a result, these parallel methods enable analog processors to provide high computational throughput and energy efficiency benefits over their conventional digital counterparts.

**Discussion**

In this section, we first discuss how to implement the SHD algorithm with 3 arrays (instead of 4) using the intuition obtained from the theoretical analysis of the coupled-system. Then we provide simulated results for a large-scale neural network for different asymmetry characteristics to benchmark our method against SGD-based training.

Considering a sequence of $m + n$ incremental and $n$ decremental changes at random order, the net modulation obtained for a symmetric device is on average $m$. On the other hand, we have shown above that for asymmetric devices the conductance value eventually converges to the symmetry point for increasing $n$ (irrespective of $m$ or the initial conductance). It can be seen by inspection that for increasing statistical variation present in the training data (causing more directional changes for updates), the effect of device asymmetry gets further pronounced, leading to heavier degradation of classification accuracy for networks trained with conventional SGD (See S3.6 and **Fig. S1**). However, this behavior can alternatively be viewed as nonlinear filtering, where only signals with persistent sign information, $\frac{m}{m+2n}$, are passed. Indeed, the SHD algorithm exploits this property within the auxiliary array, $A$, which filters the gradient information that is used to train the core array, $C$. As a result, $C$ is updated with less frequency and only in directions with a high confidence level of minimizing the error function of the problem at hand. A direct implication of this statement is that the asymmetric modulation behavior of $C$ is much less critical than that of $A$ (See S3.7 and **Fig. S2**) for successful optimization as its update signal contains less amount of statistical variation. Therefore, symmetry point information of $C_{main}$ is not relevant either. Using these results and intuition, we modified the original algorithm by discarding $C_{ref}$ and using $A_{ref}$ (set to $A_{main,symmetry}$) as a common reference array for differential readout. This modification reduces the hardware cost of SHD implementations by 50% to significantly improve their practicality.

Our description of asymmetry as the mechanism of dissipation indicates that it is a necessary and useful device property for convergence within the SHD framework (**Fig.2**E). However, this argument does not imply that the convergence speed would be determined by the magnitude of device asymmetry for practical-sized applications. Unlike the single-parameter regression problem considered above, the exploration space for DNN training is immensely large, causing optimization to take place over many iterations of the dataset. In return, the level of asymmetry required to balance (i.e. damp) the system evolution is very small and can be readily achieved by any practical level of asymmetry.

To prove these assertations, we show simulated results in **Fig. 4** for a Long Short-Term Memory (LSTM) network, using device models with increasing levels of asymmetry, trained with both the SGD and SHD algorithms. The network was trained on Leo Tolstoy's War and Peace novel, to predict the next character for a given text string (*37*). For reference, training the same network with a 32-bit digital floating-point architecture yields a cross-entropy level of 1.33 (complete learning curve shown in Fig. S6). We have particularly chosen this network as LSTM's are known for being particularly vulnerable to device asymmetry (*12*).

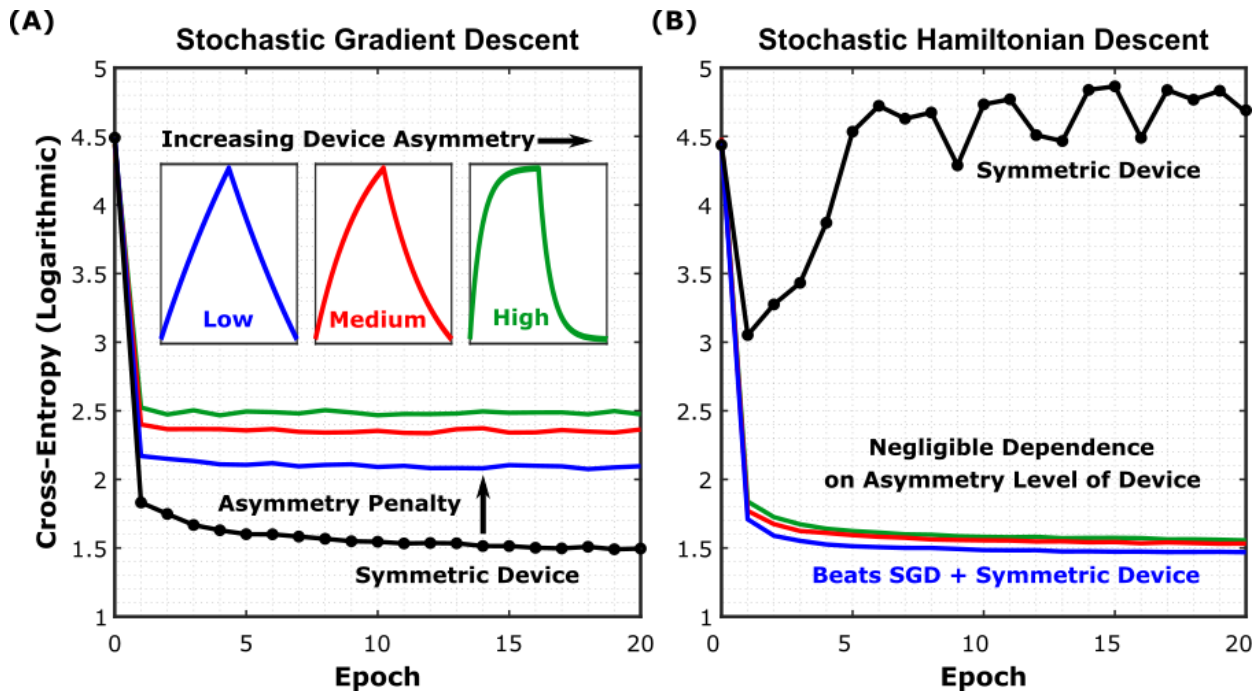

**Fig 3. Simulated training results for different resistive device technologies. (A)** Simulated learning curves of a Long Short-Term Memory (LSTM) network trained on Leo Tolstoy's War and Peace novel, using different crosspoint device models under the SGD algorithm. Details of the network can be found in Ref. (*37*) **(B)** Simulated learning curves for the same network using the SHD algorithm. All simulation details can be found in S2.7. See **Fig. S5** for device-to-device variation included in the simulations and **Fig. S6** for floating-point baseline comparison.

The insets in Fig. 4 show the average conductance modulation characteristics representative for each asymmetry level. The simulations further included device-to-device variation, cycle-to-cycle variation, analog read noise, and stochastic updating similar to the work conducted in Ref. (*9*). The learning curves show the evolution of the cross-entropy error, which measures the performance of a classification model, with respect to the epochs of training. First, **Fig 3**A shows that even for minimally asymmetric devices (blue trace) trained with SGD, the penalty in classification performance is already severe. This result also demonstrates once more the difficulty of engineering a device that is symmetric-enough to be trained accurately with SGD. On the other hand, for SHD (**Fig 3**.B), all depicted devices are trained successfully, with the sole exception being the perfectly symmetric devices (black trace), as expected (See S.3.2 and **Fig. S5** for devices with abrupt modulation characteristics). Furthermore, **Fig 3**B demonstrates that SHD can even provide training results with higher accuracy and faster convergence than those for perfectly symmetric devices trained with SGD. As a result, we conclude that SHD is generically superior to SGD for analog deep learning architectures.

Finally, although we present SHD in the context of analog computing specifically, it can also be potentially useful on conventional processors (with simulated asymmetry). The filtering dynamics described above allows SHD to guide its core component selectively in directions with high statistical persistence. Therefore, at the expense of increasing the overall memory and number of operations, SHD might outperform conventional training algorithms by providing faster convergence, better classification accuracy, and/or superior generalization performance.

**Conclusion**

In this paper, we described a fully-parallel neural network training algorithm for analog crossbar-based architectures, Stochastic Hamiltonian Descent (SHD), based on resistive devices with asymmetric conductance modulation characteristics, as is the case for all practical technologies. In contrast to previous work that resorted to serial operations to mitigate asymmetry, SHD is a fully-parallel and scalable method that can enable high throughput and energy-efficiency deep learning computations with analog hardware. Our new method uses an auxiliary array to successfully tune the system variables in order to minimize the total energy (Hamiltonian) of the system that includes the effect of device asymmetry. Standard techniques, such as Stochastic Gradient Descent, perform optimization without accounting for the effect of device asymmetry and thus converge to the minimum of a wrong energy function. Therefore, our theoretical framework describes the inherent fundamental incompatibility of asymmetric devices with conventional training algorithms. The SHD framework further enables the exploitation of device asymmetry as a useful feature to selectively filter and apply the updates only in directions with high confidence. The new insights shown here have allowed a 50% reduction in the hardware cost of the algorithm. This method is immediately applicable to a variety of existing device technologies, and complex neural network architectures, enabling the realization of analog training accelerators to tackle the ever-growing computational demand of deep learning applications.

**Acknowledgments:** We thank James B. Hannon for careful reading of our manuscript and many useful suggestions.

**Funding:** This work is funded by IBM Research.

**Author contributions:** -

**Competing interests:** The authors declare no competing interests.

**Data and materials availability:** All data are available in the main text.